\PassOptionsToPackage{numbers}{natbib}
\documentclass{article}

\usepackage[preprint]{neurips_2025}

\usepackage[utf8]{inputenc} 
\usepackage[T1]{fontenc}    
\usepackage{hyperref}       
\usepackage{url}            
\usepackage{booktabs}       
\usepackage{amsfonts}       
\usepackage{nicefrac}       
\usepackage{microtype}      
\usepackage{xcolor}         
\usepackage{graphicx}
\usepackage{import}
\usepackage{float} 
\usepackage{placeins}
\usepackage{tikz}
\usetikzlibrary{positioning, shapes, arrows.meta, fit, calc, decorations.pathreplacing,calligraphy}
\usetikzlibrary{arrows.meta, shadows.blur, backgrounds, shapes}
\usetikzlibrary{matrix}
\usepackage{pgfplots}
\usepackage{caption}
\usepackage{multirow}
\usepackage{adjustbox}
\usepackage{array}
\usepackage{cite}
\usepackage{amsmath,amssymb} 
\usepackage{makecell}
\usepackage{siunitx} \usepackage[textsize=tiny]{todonotes}
\usepackage{longtable}
\usepackage{subcaption}

\usepackage{amsmath}
\usepackage{amssymb}
\usepackage{mathtools}
\usepackage{amsthm}
\usepackage{threeparttable}

\theoremstyle{plain}

\theoremstyle{definition}

\theoremstyle{remark}

\title{Beyond Simple Concatenation: Fairly Assessing PLM Architectures for Multi-Chain Protein-Protein Interactions Prediction}

\author{
{\normalfont
Hazem Alsamkary\thanks{Equal contribution} \quad
Mohamed Elshaffei\footnotemark[1] \quad
Mohamed Soudy \quad
Sara Ossman
} \\
Abdallah Amr \quad
Nehal Adel Abdelsalam \quad
Mohamed Elkerdawy \quad
Ahmed Elnaggar\thanks{Corresponding author: publications@proteinea.com} \\
\centerline{Proteinea Inc} \\
}

\pgfplotsset{compat=1.18}
\begin{document}

\maketitle

\begin{abstract}
Protein-protein interactions (PPIs) are fundamental to numerous cellular processes, and their characterization is vital for understanding disease mechanisms and guiding drug discovery. While protein language models (PLMs) have demonstrated remarkable success in predicting protein structure and function, their application to sequence-based PPI binding affinity prediction remains relatively underexplored. This gap is often attributed to the scarcity of high-quality, rigorously refined datasets and the reliance on simple strategies for concatenating protein representations. In this work, we address these limitations. First, we introduce a meticulously curated version of the PPB-Affinity dataset of a total of 8,207 unique protein-protein interaction entries, by resolving annotation inconsistencies and duplicate entries for multi-chain protein interactions. This dataset incorporates a stringent $\leq 30\%$ sequence identity threshold to ensure robust splitting into training, validation, and test sets, minimizing data leakage. Second, we propose and systematically evaluate four architectures for adapting PLMs to PPI binding affinity prediction: embeddings concatenation (EC), sequences concatenation (SC), hierarchical pooling (HP), and pooled attention addition (PAD). These architectures were assessed using two training methods: full fine-tuning and a lightweight approach employing ConvBERT heads over frozen PLM features. Our comprehensive experiments across multiple leading PLMs (ProtT5, ESM2, Ankh, Ankh2, and ESM3) demonstrated that the HP and PAD architectures consistently outperform conventional concatenation methods, achieving up to $12\%$ increase in terms of Spearman correlation. These results highlight the necessity of sophisticated architectural designs to fully exploit the capabilities of PLMs for nuanced PPI binding affinity prediction. The code and dataset are publicly available at \url{https://github.com/Proteinea/ppiseq} and \url{https://huggingface.co/datasets/proteinea/ppb_affinity}, respectively.
\end{abstract}

\section{Introduction}
Protein-protein interactions (PPIs) are fundamental to major biological processes, underpinning complex protein networks, the assembly of functional protein structures, the operation of cellular machinery, enzymatic catalysis, the modulation of protein specificity, and signal transduction pathways \citep{braun2012history}. In translational research, the study of PPIs is crucial for identifying novel therapeutic targets and developing protein-based drugs for complex multifactorial diseases \citep{lu2020recent}. The efficacy of such protein-based therapeutics is often directly correlated with their binding affinity to their targets; consequently, achieving high binding affinity is a primary objective in drug screening and lead optimization \citep{li2009novel}. Experimental techniques are employed to quantify these interaction strengths, typically by measuring the dissociation constant ($pK_d$), a direct indicator of binding affinity. However, despite their accuracy, these experimental methods are often time-consuming, labor-intensive, resource-heavy, and characterized by low throughput \citep{mohamed2024ensembling}. This has spurred the development of computational approaches, including molecular docking and molecular dynamics simulations, aimed at predicting PPIs and their binding affinities \citep{zhao2020exploring}. Nevertheless, many existing computational methods for PPI detection and affinity quantification face challenges, including high computational costs, as well as limitations in accuracy and reliability \citep{seo2021binding}.
To overcome these limitations, machine learning (ML) techniques, including learnable scoring functions, have been implemented for PPI prediction \citep{seo2021binding}. More recently, the advent of deep learning, particularly protein language models (PLMs), has offered promising alternatives to traditional statistical learning methods. PLMs, pre-trained on vast sequence datasets \citep{uniref, bfd} using self-supervised learning, have shown remarkable success in diverse protein-related downstream tasks, including structure and function prediction \citep{esm, prottrans, esm2, elnaggar2023ankh, ankh_ext1, ankh_ext2, esm3}. This success has spurred interest in applying PLMs to the more complex task of PPI prediction, especially for quantifying binding affinity \citep{peer, plm_ppi_1, plm_ppi_2, saprot}.
However, a key challenge arises because PLMs are typically pre-trained on individual protein sequences (e.g., through masked language modeling). This design makes their native application to tasks requiring the simultaneous modeling of multiple interacting protein chains non-trivial. Consequently, current approaches encode each interacting protein independently using a PLM, then aggregate (via concatenation or addition) the fixed-length representations of all proteins before these are fed into a predictive neural network head \citep{peer, saprot}.
While straightforward to implement, these simple aggregation methods may underutilize the rich contextual information captured by PLMs. This highlights several critical gaps in current research. First, there has been limited exploration of more sophisticated architectures tailored to adapt PLMs effectively for PPI prediction. Second, the field lacks standardized techniques for handling multi-chain protein complexes, where ligands, receptors, or both may consist of multiple polypeptide chains. For instance, existing methods may concatenate all constituent chains into a single sequence, often without explicit boundary information \citep{ppi_cat1, peer}, or entirely omit multi-chain interactions from the analysis \citep{ppi_cat2}.
Furthermore, a significant impediment to progress is the absence of robust, standardized benchmarking datasets and evaluation protocols for sequence-based, multi-chain PPI binding affinity prediction. Current evaluation strategies vary widely from 10-fold cross-validation \citep{ppi_cat1} and mutation-based splits \citep{peer} to complex-level or binding-site-level leave-one-out schemes \citep{zhou2024ddmut}. This methodological heterogeneity in data pre-processing and splitting makes direct and fair comparison of different approaches difficult, thereby slowing research advancement in this domain.
To address these gaps, our research makes two primary contributions. First, we introduce a rigorously curated and standardized benchmark dataset specifically for multi-chain PPI binding affinity prediction. Derived from the recently published PPB Affinity dataset \citep{liu2024ppb}, our version features meticulous pre-processing to minimize errors. Importantly, it incorporates a strict $\leq 30\%$ sequence identity cutoff for creating training, validation, and test splits. This rigorous splitting is designed to ensure robust assessment of model generalization capabilities and minimize information leakage. The resulting dataset is made readily accessible via the Huggingface Datasets library \citep{hf_ds}, fostering standardized evaluation and promoting reproducible research.
Second, building upon this standardized benchmark, we propose and systematically evaluate novel architectures for adapting PLMs to the task of PPI binding affinity prediction. We conducted a comprehensive comparison of these proposed architectures against commonly used simple concatenation techniques. Our empirical study utilized several state-of-the-art PLMs, providing insights into both architectural efficacy and PLM-specific performance, aiming to guide future research in developing more effective computational models for PPI analysis.
\section{Methods}
\subsection{Dataset}
To train and evaluate our proposed architectures for predicting PPI binding affinity, we utilized the PPB-Affinity dataset \citep{liu2024ppb}. This dataset is a comprehensive aggregation of PPI data from multiple sources, notably including the widely recognized SKEMPI2 database \citep{skempi2}. PPB-Affinity encompasses diverse interaction types, such as antibody-antigen, TCR-pMHC, and general protein-protein interactions. Its inherent diversity provides a rich training corpus, expected to foster the development of machine learning models that generalize effectively across a broad spectrum of PPIs.
Our data curation pipeline for this dataset comprised three distinct phases:
\begin{enumerate}
    \item \textbf{Preprocessing:} This initial stage focused on identifying and correcting erroneous entries within the raw data before the extraction of protein sequences from  Protein Data Bank (PDB) files.
    \item \textbf{Postprocessing:} Following sequence extraction, this stage involved filtering and refining the protein sequences to enhance the overall data quality.
    \item \textbf{Splitting:} The final phase partitioned the curated and filtered data into distinct training, validation, and test sets for model development and robust evaluation.
\end{enumerate}
\subsubsection{Data preprocessing}
The preprocessing phase focused on enhancing data integrity by systematically identifying and rectifying or removing erroneous entries from the initial 12,062 records in the PPB-Affinity dataset.
First, we performed consistency checks for mutation annotations. Entries with incorrect mutation specifications were identified by comparing the provided mutation information against the residue in the corresponding PDB file chain. Such discrepancies can lead to incorrectly generated mutated sequences, creating a mismatch between the input protein sequences and the target binding affinity label, thereby potentially hampering model training. We identified and removed three entries due to such errors. For instance, the entry with ID '3QIW:A, B, C, D, E:C\_K8E:PMID=21490152' specified a K8E mutation in chain C; however, inspection of the PDB file revealed an 'L' at this position, not 'K'.
Second, we screened for entries referencing non-existent ligand or receptor chains in their associated PDB files. As complete protein sequence construction was impossible for these cases, such entries were excluded. We identified and removed 11 such entries. An example is the entry '5KVE:E, H, L::PMID=27475895', which specified chain 'H' as part of the ligand, although this chain was absent in the corresponding PDB file. In a related issue, several entries, predominantly associated with the SAbDab database, listed duplicate chain identifiers (e.g., 'A' and 'a') where the lowercase version did not exist in the PDB file. These entries were corrected by removing the reference to the non-existent lowercase chain. As the primary uppercase chain provided all necessary information for sequence reconstruction, these entries were therefore retained.
Finally, a specific systematic annotation error was identified: all entries associated with the PDB ID '3QIB' and originating from the ATLAS database exhibited an off-by-one error in the reported mutation site numbering for chain C. We programmatically corrected these mutation locations by incrementing the residue position by one. This ensured accurate application of mutations and maintained consistency between the derived protein sequence data and the associated binding affinity labels.
Following these preprocessing steps, the curated dataset was ready for the subsequent stage of parsing PDB files to extract protein sequences and apply the validated mutation information.

\subsubsection{Data postprocessing}
Following the initial preprocessing, the postprocessing phase further refined the dataset through several key steps to ensure data quality and suitability for training protein language models.
First, we addressed missing residues within the protein sequences derived from PDB ATOM records. The presence of such gaps, particularly non-terminal missing residues which can be integral to the protein core and thus protein stability \citep{kozlova2023proteinflow}, can significantly impact the representations generated by PLMs and potentially degrade model performance, an aspect often overlooked in literature. To ensure sequence completeness, we recovered these missing residues by referencing the corresponding sequence information available in the SEQRES records of the respective PDB files, incorporating the missing amino acids into their correct positions within the sequence. This procedure resulted in the recovery of missing residues for a total of 4,946 protein chains within the dataset.
Subsequently, we filtered out entries involving short protein chains. Specifically, any PPI entry where at least one interacting partner (ligand or receptor chain) comprised fewer than 40 residues was removed \citep{deeploc, deeploc2}. This filtering criterion was applied to ensure that our dataset predominantly contained interactions between well-defined protein domains or complete proteins, rather than protein-peptide interactions. This step resulted in the exclusion of 2,753 entries.
Finally, we addressed duplicate entries, defined as those having identical ligand and receptor sequences. Such duplicates could possess varying reported binding affinity values or be associated with different experimental measurement methods. Our consolidation strategy for these cases was as follows: if multiple affinity measurement methods were documented for a group of identical sequence pairs, entries associated with 'Unknown' or 'Other' methods were omitted to prioritize more reliable data. The binding affinity values from the remaining entries (i.e., those with explicitly known measurement methods) were then averaged to create a single, representative entry for that unique interacting pair. This de-duplication process, aimed at enhancing data quality and consistency, led to the consolidation and effective removal of 1,088 entries.
After these postprocessing steps, the final dataset prepared for model training and evaluation consisted of 8,207 unique protein-protein interaction entries.
\subsubsection{Data splitting}
Standard practices in PPI modeling often involve splitting data by randomly assigning PDB IDs to training and validation sets \citep{zhang2020mutabind2, pahari2020saambe}. Such approaches can lead to severely overestimated performance metrics due to significant information leakage between splits; for instance, random PDB ID-based splitting can result in up to 65\% leakage of training PPI interfaces into evaluation sets \citep{bushuiev2024revealing, tsishyn2024quantification}. This leakage means models are often evaluated on their ability to memorize training data rather than their capacity to generalize to genuinely unseen data. To address this critical issue, we implemented a rigorous, two-stage data splitting strategy designed to minimize information leakage and enable a fair assessment of model generalization.
The first stage involved an initial splitting of the dataset based on sequence clustering. We first created a single representative sequence for each PPI entry by concatenating all its constituent protein chains. These concatenated sequences were then clustered using the MMseqs2 \citep{mmseqs2} `cluster` algorithm, adopting a commonly used minimum sequence identity threshold of 30\% for cluster formation (i.e., between a cluster representative and any member sequence) \citep{identity_ref, deeploc, deeploc2, tape, peer, saprot, deepsol}. This process yielded 1,572 distinct clusters. An initial training split was formed by assigning the largest clusters to it until this split constituted approximately 75\% of the total dataset. This initial target of approximately 75\% was set because the subsequent refinement stage (detailed in the next paragraph) was designed to identify and reassign further qualifying sequences into the training set, thereby increasing its final proportion. Assigning larger clusters first aimed to prevent the bias of the smaller, preliminary validation and test splits towards any specific protein families. Throughout this process, we enforced PDB ID coherency, ensuring that all sequences originating from the same PDB ID were assigned to the same split.
The second stage focused on refining these initial splits to stringently control for sequence similarity across them. We iteratively used the MMseqs2 `search` command to identify any sequences remaining in the preliminary validation or test splits that exhibited a sequence identity $>30\%$ with any sequence in the formulated training split. All such identified sequences were then reassigned to the training split. During each reassignment, PDB ID coherency was maintained: if a sequence was moved, all other sequences from its parent PDB ID were also moved to the training split. This iterative refinement continued until no more sequences in the validation or test splits met the $>30\%$ similarity criterion with the training set. In total, 328 sequences were reassigned to the training data through this procedure, expanding the training split to approximately 79\% of the total dataset.
Finally, the remaining 21\% of the data was partitioned into validation and test sets. This subset was re-clustered using MMseqs2 with the identical parameters as described above, resulting in 1,071 new clusters. These clusters were then distributed as equitably as possible between the validation and test splits, ensuring that all sequences derived from the same PDB ID were allocated to only one of these final splits. This meticulous process resulted in final dataset splits of approximately 79\% for training, 12\% for validation, and 9\% for testing.

\subsection {Architecture}
Adapting PLMs for PPI binding affinity prediction involves processing multiple protein sequences, evolving from simple ligand-receptor pairs ($l,r$) to complex multi-chain entities ($Lig=(l_1,\dots,l_L)$, $Rec=(r_1,\dots,r_R)$). Since PLMs are typically pre-trained on single protein sequences \citep{prottrans, esm2, elnaggar2023ankh}, they cannot natively handle such multi-entity, variable-number inputs. Therefore, effectively leveraging PLMs for this task requires specialized architectures, which are detailed in the subsequent sections.

\subsubsection{Embeddings concatenation}
\begin{figure}[t!]
    \centering
        \begin{subfigure}[t]{0.49\textwidth}
        \begin{tikzpicture}[
            font=\footnotesize,
            >=latex,
            line join=round,
            line cap=round,
            node distance=0.3cm and 0.05cm,
            base_box/.style={
                draw,
                rounded corners,
                inner sep=1.0pt,
                minimum width=0.0cm,
                minimum height=0.1cm,
                align=center
            },
            input_box/.style={ 
                base_box,
                shape=rectangle,
                rotate=0,
                fill=gray!15
            },
            plm_box/.style={ 
                base_box,
                fill=red!30
            },
            pool1_box/.style={ 
                base_box,
                fill=yellow!30
            },
            pool2_box/.style={ 
                base_box,
                fill=yellow!30
            },
            concat_box/.style={ 
                base_box,
                fill=orange!30
            },
            mlp_box/.style={ 
                base_box,
                fill=violet!30
            }
        ]
        
        
        \node[input_box] (lig_in) {\begin{tabular}{c}
        \(l_1\text{[EOS]}l_2\dots l_L\text{[EOS]}\) \\ 
        Ligand Chains 
        \end{tabular}};
        
        \node[input_box, right=of lig_in] (rec_in) {\begin{tabular}{c}
        \(r_1\text{[EOS]}r_2\dots r_R\text{[EOS]}\) \\ 
        Receptor Chains 
        \end{tabular}};
        
        
        \node[plm_box, above=of lig_in] (lig_plm) {
        \begin{tabular}{c}
        \(\rightarrow (\text{len}_{lig}\times E_{dim})\) \\ 
        PLM 
        \end{tabular}
        };
        \draw[->, thick] (lig_in) -- (lig_plm);
        
        \node[pool2_box, above=of lig_plm] (lig_pool_chain) {
        \begin{tabular}{c}
        \(\rightarrow (E_{\dim})\)\\ 
        Attention Pooler 
        \end{tabular}
        };
        \draw[->, thick] (lig_plm) -- (lig_pool_chain);

        \node[plm_box, above=of rec_in] (rec_plm) {
        \begin{tabular}{c}
        \(\rightarrow (\text{len}_{rec}\times E_{dim})\)  \\ 
        PLM 
        \end{tabular}
        };
        \draw[->, thick] (rec_in) -- (rec_plm);
        
        \node[pool2_box, above=of rec_plm] (rec_pool_chain) {
        \begin{tabular}{c}
        \(\rightarrow (E_{\dim})\)\\ 
        Attention Pooler 
        \end{tabular}
        };
        \draw[->, thick] (rec_plm) -- (rec_pool_chain);

        
        \coordinate (mid_point) at ($(lig_pool_chain)!0.5!(rec_pool_chain)$);
        
        \node[concat_box, above=of mid_point, yshift=0.8cm] (concat) {
        \begin{tabular}{c}
        \(\rightarrow (2\,E_{\dim})\)\\ 
        Concatenate 
        \end{tabular}
        };
        
        \draw[->, thick] (lig_pool_chain.north) -- ++(0,0.4cm) coordinate (lc_inter)
            -- (lc_inter -| concat.south west)
            -- (concat.south west);
        
        \draw[->, thick] (rec_pool_chain.north) -- ++(0,0.4cm) coordinate (rc_inter)
            -- (rc_inter -| concat.south east)
            -- (concat.south east);
        
        \node[mlp_box, above=of concat] (mlp) {
        \begin{tabular}{c}
        \(\rightarrow \mathbb{R}^1\)\\ 
        MLP 
        \end{tabular}
        };
        \draw[->, thick] (concat) -- (mlp);
        
        \end{tikzpicture}
        \caption{Embeddings Concatenation (EC) Architecture}
    \label{fig:ec}
\end{subfigure}%
        ~
        \begin{subfigure}[t]{0.49\textwidth}
    \centering
        \begin{tikzpicture}[
            font=\footnotesize,
            >=latex,
            line join=round,
            line cap=round,
            node distance=0.3cm and 0.97cm,
            base_box/.style={
                draw,
                rounded corners,
                inner sep=1.0pt,
                minimum width=0.0cm,
                minimum height=0.1cm,
                align=center
            },
            input_box/.style={ 
                base_box,
                shape=rectangle,
                rotate=0,
                fill=gray!15
            },
            plm_box/.style={ 
                base_box,
                fill=red!30
            },
            pool1_box/.style={ 
                base_box,
                fill=yellow!30
            },
            pool2_box/.style={ 
                base_box,
                fill=yellow!30
            },
            concat_box/.style={ 
                base_box,
                fill=orange!30
            },
            mlp_box/.style={ 
                base_box,
                fill=violet!30
            }
        ]
        
        
        \node[input_box] (lig_in) {\begin{tabular}{c}
        \((l_1,l_2,\dots,l_L)\)\\ 
        Ligand Chains 
        \end{tabular}};
        
        \node[input_box, right=of lig_in] (rec_in) {\begin{tabular}{c}
        \((r_1,r_2,\dots,r_R)\)\\  
        Receptor Chains 
        \end{tabular}};
        
        
        \node[plm_box, above=of lig_in] (lig_plm) {
        \begin{tabular}{c}
        \(\rightarrow \{(\text{len\_}l_1\times E_{dim}),\)\\\(\dots\),\\\((\text{len\_}l_L\times E_{dim})\}\) \\ 
        PLM 
        \end{tabular}
        };
        \draw[->, thick] (lig_in) -- (lig_plm);
        
        \node[pool1_box, above=of lig_plm] (lig_pool_seq) {
        \begin{tabular}{c}
        \(\rightarrow (L\times E_{\dim})\)\\ 
        Attention Pooler 
        \end{tabular}
        };
        \draw[->, thick] (lig_plm) -- (lig_pool_seq);
        
        \node[pool2_box, above=of lig_pool_seq] (lig_pool_chain) {
        \begin{tabular}{c}
        \(\rightarrow (E_{\dim})\)\\ 
        Attention Pooler 
        \end{tabular}
        };
        \draw[->, thick] (lig_pool_seq) -- (lig_pool_chain);

        \node[plm_box, above=of rec_in] (rec_plm) {
        \begin{tabular}{c}
        \(\rightarrow \{(\text{len\_}r_1\times E_{dim}),\)\\\(\dots\)\\,\((\text{len\_}r_R\times E_{dim})\}\) \\ 
        PLM 
        \end{tabular}
        };
        \draw[->, thick] (rec_in) -- (rec_plm);
        
        \node[pool1_box, above=of rec_plm] (rec_pool_seq) {
        \begin{tabular}{c}
        \(\rightarrow (R\times E_{\dim})\)\\ 
        Attention Pooler 
        \end{tabular}
        };
        \draw[->, thick] (rec_plm) -- (rec_pool_seq);
        
        \node[pool2_box, above=of rec_pool_seq] (rec_pool_chain) {
        \begin{tabular}{c}
        \(\rightarrow (E_{\dim})\)\\ 
        Attention Pooler 
        \end{tabular}
        };
        \draw[->, thick] (rec_pool_seq) -- (rec_pool_chain);

        
        \coordinate (mid_point) at ($(lig_pool_chain)!0.5!(rec_pool_chain)$);
        
        \node[concat_box, above=of mid_point, yshift=0.8cm] (concat) {
        \begin{tabular}{c}
        \(\rightarrow (2\,E_{\dim})\)\\ 
        Concatenate 
        \end{tabular}
        };
        
        \draw[->, thick] (lig_pool_chain.north) -- ++(0,0.4cm) coordinate (lc_inter)
            -- (lc_inter -| concat.south west)
            -- (concat.south west);
        
        \draw[->, thick] (rec_pool_chain.north) -- ++(0,0.4cm) coordinate (rc_inter)
            -- (rc_inter -| concat.south east)
            -- (concat.south east);
        
        \node[mlp_box, above=of concat] (mlp) {
        \begin{tabular}{c}
        \(\rightarrow \mathbb{R}^1\)\\ 
        MLP 
        \end{tabular}
        };
        \draw[->, thick] (concat) -- (mlp);
        
        \end{tikzpicture}
    \caption{Hierarchical Pooling (HP) Architecture}
    \label{fig:hp}
\end{subfigure}
        ~
        \begin{subfigure}[t]{0.45\textwidth}
    \centering
        \begin{tikzpicture}[
            font=\footnotesize,
            >=latex,
            line join=round,
            line cap=round,
            node distance=0.3cm and 0.5cm,
            base_box/.style={
                draw,
                rounded corners,
                inner sep=1.0pt,
                minimum width=0.0cm,
                minimum height=0.1cm,
                align=center
            },
            input_box/.style={ 
                base_box,
                shape=rectangle,
                rotate=0,
                fill=gray!15
            },
            plm_box/.style={ 
                base_box,
                fill=red!30
            },
            pool1_box/.style={ 
                base_box,
                fill=yellow!30
            },
            pool2_box/.style={ 
                base_box,
                fill=yellow!30
            },
            concat_box/.style={ 
                base_box,
                fill=orange!30
            },
            attn_box/.style={ 
                base_box,
                fill=cyan!30
            },
            mlp_box/.style={ 
                base_box,
                fill=violet!30
            },
            sum_node/.style={circle, draw, inner sep=1pt, minimum size=3mm} 
        ]
        
        
        \node[input_box] (lig_in) {\begin{tabular}{c}
        \(l_1\text{[EOS]}l_2\dots l_L\text{[EOS]}\) \\ 
        Ligand Chains 
        \end{tabular}};
        
        \node[input_box, right=of lig_in] (rec_in) {\begin{tabular}{c}
        \(r_1\text{[EOS]}r_2\dots r_R\text{[EOS]}\) \\ 
        Receptor Chains 
        \end{tabular}};
        
        
        \node[plm_box, above=of lig_in] (lig_plm) {
        \begin{tabular}{c}
        \(\rightarrow (\text{len}_{lig}\times E_{dim})\) \\ 
        PLM 
        \end{tabular}
        };
        \draw[->, thick] (lig_in) -- (lig_plm);

        \node[attn_box, above=of lig_plm] (lig_attn) {
        \begin{tabular}{c}
        \(\rightarrow (\text{len}_{lig}\times E_{dim})\) \\ 
        Multi-Head\\Cross-Attention 
        \end{tabular}
        };
        \draw[->, thick] (lig_plm) -- (lig_attn);

        \node[sum_node, above=0.5cm of lig_attn] (lig_sum) {$+$}; 
    
        \draw[->, thick] (lig_attn) -- (lig_sum);
    
        \coordinate (lig_res_h) at ($(lig_plm.west)-(0.3cm,0)$); 
        \coordinate (lig_res_v) at (lig_res_h |- lig_sum.west);        
        \draw[->, thick] (lig_plm.west) -- (lig_res_h) -- (lig_res_v) -- (lig_sum.west); 
        
        \node[pool2_box, above=of lig_sum] (lig_pool_chain) {
        \begin{tabular}{c}
        \(\rightarrow (E_{\dim})\)\\ 
        Attention Pooler 
        \end{tabular}
        };
        \draw[->, thick] (lig_sum) -- (lig_pool_chain);

        \node[plm_box, above=of rec_in] (rec_plm) {
        \begin{tabular}{c}
        \(\rightarrow (\text{len}_{rec}\times E_{dim})\)  \\ 
        PLM 
        \end{tabular}
        };
        \draw[->, thick] (rec_in) -- (rec_plm);

        \node[attn_box, above=of rec_plm] (rec_attn) {
        \begin{tabular}{c}
        \(\rightarrow (\text{len}_{rec}\times E_{dim})\) \\ 
        Multi-Head\\Cross-Attention 
        \end{tabular}
        };
        \draw[->, thick] (rec_plm) -- (rec_attn);
        \draw[->, thick] (rec_plm) -- (lig_attn);
        \draw[->, thick] (lig_plm) -- (rec_attn);

        \node[sum_node, above=0.5cm of rec_attn] (rec_sum) {$+$}; 
    
        \draw[->, thick] (rec_attn) -- (rec_sum);
    
        \coordinate (rec_res_h) at ($(rec_plm.east)+(0.3cm,0)$); 
        \coordinate (rec_res_v) at (rec_res_h |- rec_sum.east);        
        \draw[->, thick] (rec_plm.east) -- (rec_res_h) -- (rec_res_v) -- (rec_sum.east); 
        
        \node[pool2_box, above=of rec_sum] (rec_pool_chain) {
        \begin{tabular}{c}
        \(\rightarrow (E_{\dim})\)\\ 
        Attention Pooler 
        \end{tabular}
        };
        \draw[->, thick] (rec_sum) -- (rec_pool_chain);

        
        \coordinate (mid_point) at ($(lig_pool_chain)!0.5!(rec_pool_chain)$);
        
        \node[sum_node, above=of mid_point, yshift=0.8cm] (final_sum) {$+$};
        
        \draw[->, thick] (lig_pool_chain.north) |- (final_sum.west);
        
        \draw[->, thick] (rec_pool_chain.north) |- (final_sum.east);
        
        \node[mlp_box, above=of final_sum] (mlp) {
        \begin{tabular}{c}
        \(\rightarrow \mathbb{R}^1\)\\ 
        MLP 
        \end{tabular}
        };
        \draw[->, thick] (final_sum) -- (mlp);
        
        \end{tikzpicture}
    \caption{Pooled Attention Addition (PAD) Architecture}
    \label{fig:pad}
\end{subfigure}
\caption{\small{Architectures used to adapt protein language models to the binding-affinity prediction task. Dimensions displayed on each block denote component output dimensions; all weights were shared across parallel ligand and receptor processing pathways}}
\end{figure}
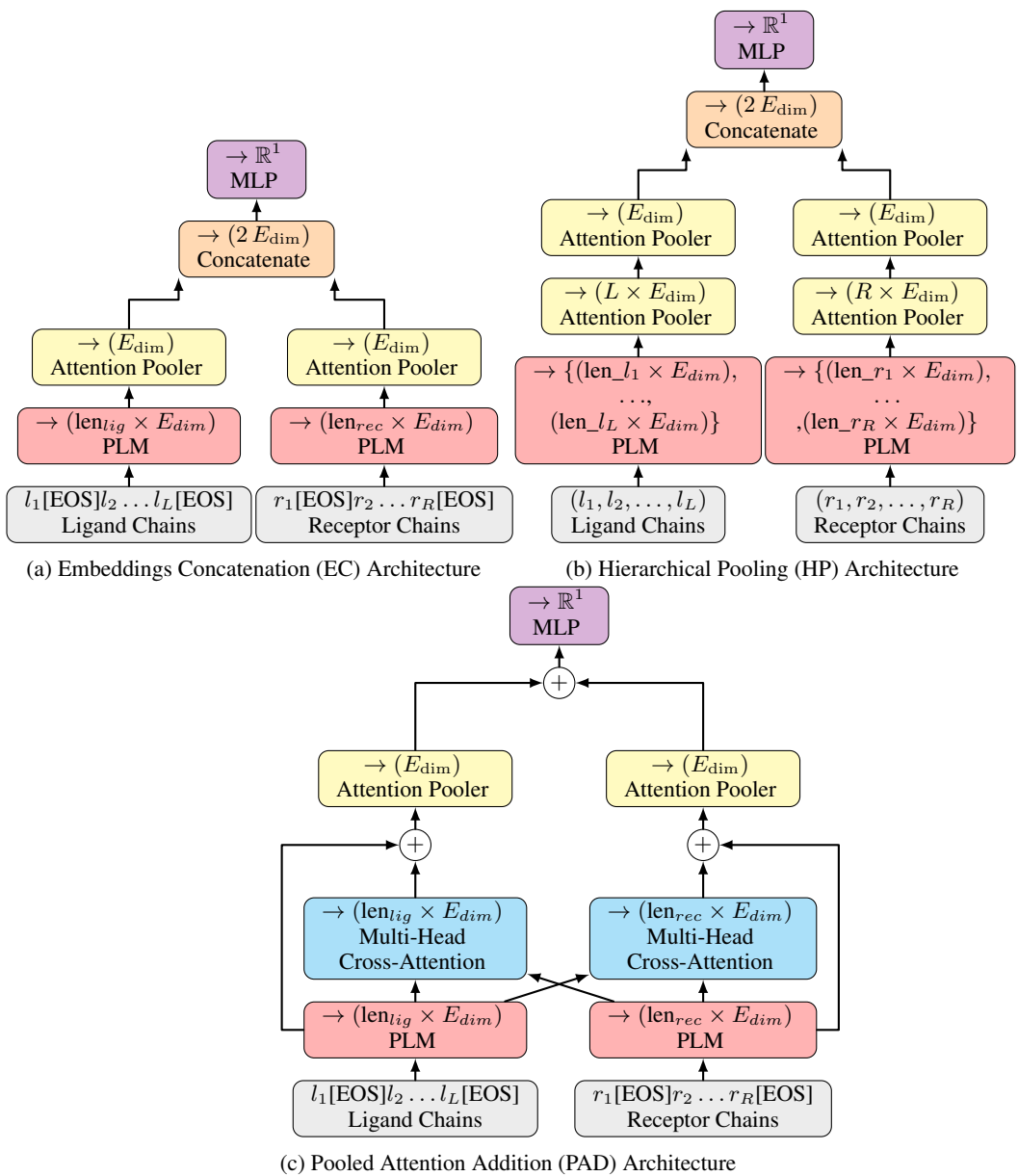

The embeddings concatenation (EC) architecture (Figure \ref{fig:ec}) adapts PLMs to PPI tasks \citep{peer, saprot}. For each interacting partner (ligand or receptor), its constituent chains are concatenated with inter-chain End-of-Sequence (EOS) tokens for boundary demarcation. This composite sequence is then encoded by a PLM (outputting embeddings of dimension $E_{dim}$) into a sequence of hidden states with dimensions $(\text{len}_{\text{partner}} \times E_{dim})$. These hidden states are subsequently condensed by a 1D global attention pooling layer (supplementary Figure \ref{fig:attn_poolin}) to produce a fixed-dimension $E_{dim}$ embedding for that partner. Both the PLM and the attention pooler share trainable weights across these parallel ligand and receptor processing pathways. Finally, the resulting $E_{dim}$ ligand and receptor embeddings are concatenated (forming a combined vector of dimension $2E_{dim}$) and input to a multi-layer perceptron (MLP) for binding affinity prediction.
\subsubsection{Sequences concatenation}
The sequences concatenation (SC) architecture (supplementary Figure \ref{fig:sc}) adapts a PLM to a PPI task by concatenating the input protein sequences before they are processed by the PLM, in contrast to EC architecture which concatenates PLM output representations.
In the SC approach, all individual protein chains from both the ligand and the receptor are combined to form a single, unified input sequence. EOS tokens are inserted between each original chain sequence within this composite input to mark their respective boundaries. This entire concatenated sequence is then fed into the PLM. The PLM processes this sequence to produce a corresponding sequence of hidden state representations, capturing contextual information across all constituent chains. These representations typically have dimensions of $(\text{len}_{\text{concat}} \times E_{dim})$, where $\text{len}_{\text{concat}}$ is the total length of the concatenated input sequence and $E_{dim}$ is the embedding dimension of the PLM.
Subsequently, this sequence of hidden state representations is passed to a 1D global attention pooling layer. This pooling layer aggregates information across the entire sequence of hidden states, outputting a single, fixed-dimension embedding of $E_{dim}$. Finally, this comprehensive embedding is fed into a MLP, which serves as the prediction head, to map it to a single binding affinity value.
\subsubsection{Hierarchical pooling}
Concatenating multiple input protein chains, as performed in EC and SC architectures, can lead to a significant GPU memory footprint, particularly when dealing with long sequences or complexes with many chains. To address this limitation, we propose the hierarchical pooling (HP) architecture (Figure \ref{fig:hp}). Unlike EC and SC, HP avoids the concatenation of input sequences to the PLM.
The HP architecture processes interacting partners (ligand and receptor) as follows. Considering the ligand, which may consist of $L$ individual chains ($l_1, l_2, \dots, l_L$):
\begin{enumerate}
    \item \textbf{Individual chain encoding:} Each of the $L$ ligand chains is fed independently into the PLM. This produces $L$ distinct sequences of hidden state representations, where each sequence corresponds to an individual chain and its dimensions are (sequence length $\times E_{dim}$).
    \item \textbf{Intra-chain pooling (First-level pooling):} Each of these $L$ hidden state sequences is then independently processed by a 1D global attention pooling layer. This step generates $L$ fixed-dimension chain embeddings with output shape of $(L,E_{dim})$, effectively summarizing the information from each individual ligand chain.
    \item \textbf{Inter-chain pooling (Second-level pooling):} The $L$ resulting chain embeddings, which form an input tensor of shape $(L, E_{dim})$, are subsequently pooled together using another 1D global attention pooling layer. This "pooling across chains" step aggregates information across the chain dimension (i.e., across the $L$ chains), collapsing the input into a single, comprehensive ligand-level representation of dimension $E_{dim}$.
\end{enumerate}
An identical hierarchical procedure is carried out for the $R$ receptor chains to compute a corresponding receptor-level representation of dimension $E_{dim}$. All trainable weights within the PLM and both levels of attention pooling layers are shared between the parallel ligand and receptor processing pathways.
Finally, the computed ligand-level and receptor-level embeddings are concatenated, forming a single feature vector of dimension $2E_{dim}$. This combined vector is then fed into a MLP prediction head, which maps it to the final binding affinity value.
\subsubsection{Pooled attention addition}
To introduce a more explicit inductive bias for learning direct ligand-receptor interactions, which may be less emphasized in architectures relying solely on various forms of concatenation, we propose the pooled attention addition (PAD) architecture (Figure \ref{fig:pad}).
The initial processing stages of the PAD architecture are similar to those in the EC approach. For the ligand, all its constituent chains are concatenated into a single sequence, with EOS tokens inserted between chains to demarcate their boundaries. This composite sequence is then processed by a PLM to generate a sequence of hidden states, $E_{lig}$, with dimensions $(\text{len}_{lig} \times E_{dim})$, where $\text{len}_{lig}$ is the length of the concatenated ligand sequence and $E_{dim}$ is the PLM's embedding dimension. An analogous procedure, employing a PLM with shared weights, is used for the receptor chains to compute the receptor's hidden state sequence, $E_{rec}$, with dimensions $(\text{len}_{rec} \times E_{dim})$.
The core of the PAD architecture lies in its interaction modeling through a Multi-head attention (MHA) block \citep{vaswani2017attention}, which computes cross-attention between the ligand and receptor hidden state sequences. Specifically, the receptor-contextualized ligand representation, $a_{lig}$, is computed by feeding $E_{lig}$ as the Query (Q) sequence, and $E_{rec}$ as both the Key (K) and Value (V) sequences to the MHA block. Symmetrically, the ligand-contextualized receptor representation, $a_{rec}$, is then computed by using $E_{rec}$ as the Q sequence, and $E_{lig}$ as both K and V sequences. This cross-attention mechanism is designed to explicitly model the inter-dependencies between the ligand and receptor representations. The resulting attention outputs are then integrated back into their respective original sequences via element-wise addition using residual connections: $E'_{lig} = E_{lig} + a_{lig}$ and $E'_{rec} = E_{rec} + a_{rec}$.
Subsequently, these updated hidden state sequences, $E'_{lig}$ and $E'_{rec}$, are independently processed by 1D global attention pooling layers. This step yields a fixed-dimension embedding for the ligand and a fixed-dimension embedding for the receptor, each of dimension $E_{dim}$. Unlike other architectures that concatenate these embeddings, the PAD architecture combines them through element-wise vector addition. The resulting sum vector, also of dimension $E_{dim}$, is then passed to a MLP prediction head to output the final binding affinity value. All trainable weights in the parallel PLM encoding, MHA block, and 1D global attention pooling stages for the ligand and receptor paths are shared.
\subsection {Training details}
We evaluated two primary training paradigms: (1) full fine-tuning of the PLM, adapting all its parameters, which is computationally intensive but may yield higher performance; and (2) a lightweight approach training a downstream head on features extracted from a frozen PLM. For the latter, we employed a ConvBERT model \citep{convbert} as the downstream head, selected for its previously reported state-of-the-art efficacy \citep{elnaggar2023ankh}. This feature-extraction strategy significantly reduces memory demands, a crucial factor for architectures like EC, SC, and PAD that involve input sequences concatenation, leading to longer sequence lengths for the PLM. These training paradigms were applied across a suite of PLMs: ProtT5 \citep{prottrans}, ESM2 (650M and 3B variants) \citep{esm2}, Ankh (Base and Large variants) \citep{elnaggar2023ankh}, Ankh2 (Ext1 and Ext2 variants) \citep{ankh_ext1, ankh_ext2}, and ESM3-SM-Open \citep{esm3}.
All models were trained by minimizing the mean squared error (MSE) loss between the predicted and ground-truth binding affinities (measured in $pK_d$). Full PLM fine-tuning utilized brain float16 (bf16) mixed precision, whereas the ConvBERT head was trained with float32 precision. The ConvBERT head consisted of a single layer with an intermediate size of $0.5 \times E_{dim}$ (where $E_{dim}$ is the PLM's embedding dimension), a kernel size of 7, a hidden dropout probability of 0.2, and an attention dropout probability of 0.1. For the EC, HP, and PAD architectures, ConvBERT weights were shared between the ligand and receptor processing pathways. Training proceeded for a maximum of 30 epochs, with early stopping if the Spearman correlation coefficient ($\rho$) on the validation set did not improve for five consecutive epochs. The checkpoint achieving the highest validation $\rho$ was selected for final evaluation. All reported metrics were the mean values from three independent runs using different random seeds.
Standard hyperparameters (detailed in Table \ref{tab:hyperparams}) were used for most experiments. Exceptions were made for full fine-tuning of ESM2-3B and ESM3-SM-Open, which required a reduced learning rate for stable convergence. Additionally, for ESM3-SM-Open under full fine-tuning, per-residue representations from the PLM were L2-normalized to mitigate training instability caused by large magnitude outputs. All models were trained using PyTorch \citep{pytorch} Distributed Data Parallel across two GPUs (either NVIDIA H100 or A6000, yielding identical results), leveraging available hardware to expedite experimentation.

\begin{table}[h!] 
\centering 
\caption{Training hyperparameters for full finetuning and ConvBERT training across all architectures} 
\label{tab:hyperparams} 
    \begin{tabular}{ll}
    \toprule 
    \textbf{Hyperparameter} & \textbf{Value} \\ 
    \midrule 
    Optimizer & Adam \\
    Learning Rate & \makecell[l]{$5 \times 10^{-4}$ \\ ($5 \times 10^{-5}$ fine-tuning ESM2-3B and ESM3)} \\ 
    Learning Rate Scheduler & Linear with warmup \\
    Warmup Steps & 1000 \\
    Number of GPUs & 2 \\ 
    Per Device Batch Size & 1 \\
    Gradient Accumulation Steps & 32 \\
    Total Effective Batch Size & 64 (1 $\times$ 32 $\times$ 2) \\ 
    Random Seeds & 7, 8, 9 \\
    \bottomrule 
    \end{tabular}
\end{table}
\section{Results and discussion}
Analysis of the top 10 performing runs, ranked by Spearman correlation coefficient on the test set (Table \ref{tab:results} highlights key trends in adapting PLMs for binding affinity prediction (The results of all the runs are available in Supplementary Table \ref{tab_full}). While the single highest correlation was achieved by ProtT5 using the PAD architecture with a ConvBERT head, the top three entries featured distinct combinations of PLMs, training methods, and architectures. This diversity suggests that optimal performance is not confined to a singular configuration.\\ 
\begin{table*}[tp]
\resizebox{\textwidth}{!}{
\centering
\begin{threeparttable}
\caption{Top 10 runs ranked by test set Spearman correlation ($\rho$). Metrics (mean $\pm$ standard deviation, 3 seeds): Spearman $\rho$, Pearson r, and RMSE ($pK_d$). PAD: Pooled attention addition; HP:
Hierarchical pooling}
\label{tab:results}
    \begin{tabular}{ll ccc ccc}
    \toprule
    \multirow{2}{*}{\textbf{PLM}} & \multirow{2}{*}{\textbf{Setup}} &
    \multicolumn{3}{c}{\textbf{Validation Split}} & \multicolumn{3}{c}{\textbf{Test Split}} \\
    \cmidrule(lr){3-5}\cmidrule(lr){6-8}
     & & \textbf{Spearman} & \textbf{Pearson} & \textbf{RMSE} & \textbf{Spearman} & \textbf{Pearson} & \textbf{RMSE} \\
    \midrule
    Prot-T5 & ConvBERT-PAD
      & \textbf{0.48 $\pm$ 0.02} & 0.48 $\pm$ 0.01 & 1.52 $\pm$ 0.09
      & \textbf{0.48 $\pm$ 0.03} & \textbf{0.51 $\pm$ 0.02} & 1.42 $\pm$ 0.10 \\
    
    Ankh2-Ext1 & Finetuning-HP
      & \textbf{0.48 $\pm$ 0.01} & \textbf{0.49 $\pm$ 0.01} & 1.50 $\pm$ 0.10
      & 0.47 $\pm$ 0.01 & 0.48 $\pm$ 0.01 & 1.45 $\pm$ 0.13 \\
    
    ESM2-650M & ConvBERT-HP
      & 0.44 $\pm$ 0.03 & 0.43 $\pm$ 0.03 & 1.86 $\pm$ 0.29
      & 0.47 $\pm$ 0.02 & 0.48 $\pm$ 0.02 & 1.74 $\pm$ 0.33 \\
    
    Ankh2-Ext2 & Finetuning-HP
      & 0.47 $\pm$ 0.01 & 0.47 $\pm$ 0.01 & 1.51 $\pm$ 0.07
      & 0.45 $\pm$ 0.01 & 0.46 $\pm$ 0.02 & 1.43 $\pm$ 0.06 \\
    
    ESM2-650M & Finetuning-HP
      & 0.45 $\pm$ 0.02 & 0.44 $\pm$ 0.02 & 1.75 $\pm$ 0.29
      & 0.44 $\pm$ 0.02 & 0.45 $\pm$ 0.01 & 1.68 $\pm$ 0.29 \\
    
    Ankh-Base & Finetuning-HP
      & 0.47 $\pm$ 0.01 & 0.47 $\pm$ 0.00 & \textbf{1.47 $\pm$ 0.03}
      & 0.44 $\pm$ 0.01 & 0.45 $\pm$ 0.01 & \textbf{1.41 $\pm$ 0.03} \\
    
    Prot-T5 & ConvBERT-HP
      & 0.42 $\pm$ 0.02 & 0.42 $\pm$ 0.02 & 1.66 $\pm$ 0.23
      & 0.44 $\pm$ 0.01 & 0.44 $\pm$ 0.01 & 1.59 $\pm$ 0.25 \\
    
    Prot-T5 & Finetuning-PAD
      & 0.44 $\pm$ 0.00 & 0.44 $\pm$ 0.00 & 1.65 $\pm$ 0.13
      & 0.44 $\pm$ 0.01 & 0.45 $\pm$ 0.01 & 1.57 $\pm$ 0.14 \\
    
    Prot-T5 & Finetuning-HP
      & 0.47 $\pm$ 0.02 & 0.46 $\pm$ 0.02 & 1.51 $\pm$ 0.05
      & 0.44 $\pm$ 0.01 & 0.45 $\pm$ 0.01 & 1.47 $\pm$ 0.08 \\
    
    ESM2-3B & Finetuning-PAD
      & 0.45 $\pm$ 0.01 & 0.45 $\pm$ 0.01 & 1.49 $\pm$ 0.07
      & 0.44 $\pm$ 0.01 & 0.46 $\pm$ 0.00 & 1.43 $\pm$ 0.08 \\
    \bottomrule
    \end{tabular}    
    \end{threeparttable}
}
\end{table*}

\begin{figure}[!ht]
    \centering
    \includegraphics[width=\textwidth]{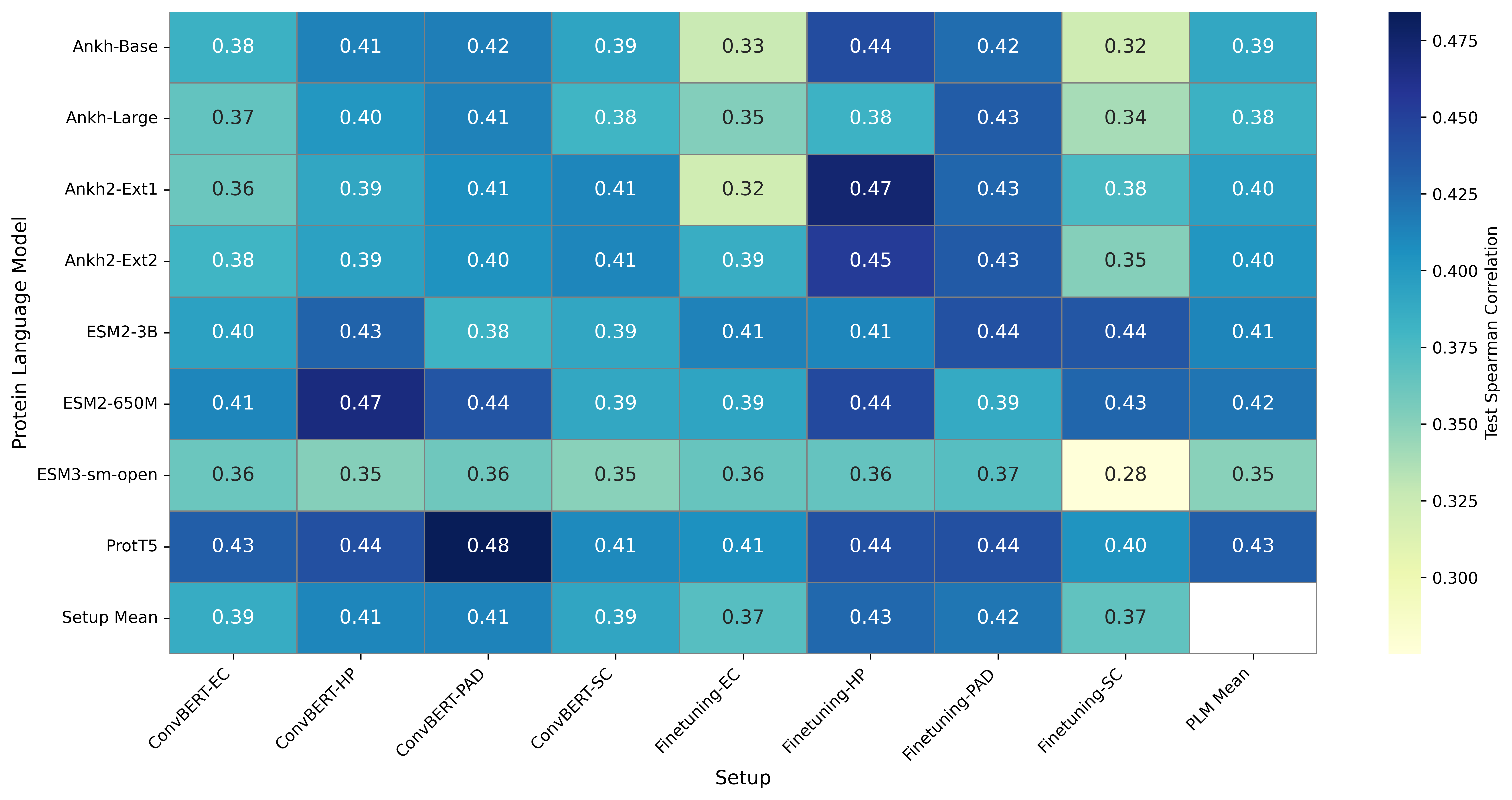}
    \caption{\small{Heatmap of test set Spearman $\rho$ (each value averaged over 3 seeds): PLMs vs. setups for binding affinity prediction. Marginal means show average $\rho$ per PLM (last column) and per setup (last row)}. PAD: Pooled Attention Addition; HP: Hierarchical Pooling; SC: Sequences Concatenation; EC: Embeddings Concatenation}
    \label{fig:boxplot_combs}
\end{figure}

Notably, the HP architecture prevailed in 7 of the top 10 runs, and PAD also demonstrated consistently strong performance. On the other hand, the simpler EC and SC architectures were conspicuously absent from this top tier, indicating their limitations for this complex task.\\
The overall performance (Figure \ref{fig:boxplot_combs}) indicated by test Spearman correlations for all PLMs across all setups (defined as a training method-architecture combination), underscores the critical role of architectural design. The HP and PAD architectures significantly outperformed EC and SC, particularly when employing full PLM fine-tuning.
The marked underperformance of SC may stem from its approach of concatenating all ligand and receptor chains into one input sequence. This strategy can obscure the distinction between ligand and receptor entities for the PLM, as the same EOS token separates all chains. Furthermore, the PLM's self-attention mechanism, when applied to such a long, composite sequence, might not effectively differentiate crucial inter-chain interactions from intra-chain relationships, leading to a dilution of the interaction signal that is potentially aggravated by the final global pooling step.\\

The superiority of HP over EC, despite their identical behavior for single-chain ligand/receptor pairs, is particularly instructive and points to more effective handling of multi-chain complexes by HP. In the EC architecture, concatenating all chains of a partner (e.g., ligand) before PLM encoding and subsequently pooling this combined, lengthy representation can dilute chain-specific features and lead to information loss during the compression to a fixed-size vector. In contrast, the hierarchical strategy of HP processes each chain individually through the PLM, performs chain-level attention pooling to capture distinct characteristics, and only then aggregates these rich chain embeddings into a partner-level representation. This preserves vital chain-level information more effectively, providing a more robust foundation for predicting interactions in multi-chain scenarios.\\
The PAD architecture also demonstrated superior performance compared to EC, despite sharing a similar initial chain concatenation strategy for forming ligand and receptor representations. This advantage can be attributed to its integrated cross-attention block. This mechanism introduces a strong inductive bias for modeling direct ligand-receptor interdependencies, generating refined, context-aware representations for both partners before pooling. These refined representations are likely less susceptible to information loss during the subsequent pooling and final aggregation steps (element-wise addition in PAD), contributing to its enhanced predictive power over EC and SC.\\
No single PLM consistently outperformed others across all setups; rather, PLM efficacy was highly dependent on the chosen architecture and training strategy (Figure \ref{fig:boxplot_combs}). ProtT5, for instance, achieved the highest average Spearman correlation, demonstrating robust performance across various configurations. However, in specific setups, other models excelled; for example, ESM2-650M and Ankh2-Ext1 showed superior results when combined with the HP architecture.
Conversely, ESM3-SM-Open generally underperformed. During full fine-tuning, it exhibited training instability, characterized by large hidden state values, which persisted to some degree despite mitigation like reduced learning rates and L2 normalization of representations. While training was stable when using ESM3-SM-Open with ConvBERT heads (frozen PLM), its performance remained comparatively low. This suggests that ESM3-SM-Open might require more extensive, task-specific hyperparameter tuning or possess inherent characteristics less suited to this PPI binding affinity task within the evaluated configurations. Notably, model size was not a definitive predictor of performance, as exemplified by ESM2-650M outperforming the larger ESM2-3B model both overall and in multiple specific setups.\\
The choice between full PLM fine-tuning and using a lightweight ConvBERT head with a frozen PLM was also dependent on the specific PLM (Figure \ref{fig:boxplot_combs}). For ESM2 models, ESM3-SM-Open, and ProtT5, full fine-tuning offered minimal, if any, performance gains over the ConvBERT approach. In such cases, the substantially lower computational and memory requirements of the ConvBERT strategy present a more efficient alternative. In contrast, for the Ankh and Ankh2 PLMs, full fine-tuning, particularly when paired with the HP or PAD architectures, generally yielded significantly better results than the ConvBERT-based feature extraction approach. This indicates that for certain PLMs, allowing the model's deeper layers to adapt to the downstream task is crucial for unlocking their full potential in PPI binding affinity prediction, especially with sophisticated interaction architectures.
\section{Conclusion and future work}
This work underscores the necessity of meticulously processed datasets with stringent, low-leakage splits for the reliable evaluation of PLMs in PPI binding affinity prediction. We demonstrated that optimal model performance is not solely dependent on PLM choice but is also determined by the interplay between the PLM, the chosen training strategy, and the architecture used. Our findings reveal that the proposed hierarchical pooling and pooled attention addition architectures significantly surpass commonly used concatenation-based approaches, offering more effective strategies for leveraging PLMs in this vital domain and expanding the horizons for their application.
We acknowledge limitations stemming from computational constraints, which prevented extensive, PLM-specific hyperparameter tuning that may potentially boost performance. Our analysis was also limited to sequence-based prediction. Incorporating protein structure, particularly relevant for models like ESM3, was not feasible due to inconsistencies created during missing residue recovery in sequences compared to original PDB structures, and the prohibitive computational cost of predicting updated structures for all affected entries. Future directions include dedicated hyperparameter optimization for individual PLMs and integrating predicted 3D structural information. Using tools like AlphaFold \citep{alphafold} to generate structures for processed sequences would facilitate developing and testing multi-modal models, potentially enhancing PLMs capable of leveraging structural features.

\FloatBarrier

\begin{ack}
The authors gratefully acknowledge the collaborative efforts and significant contributions of the Proteinea team. We especially thank Ahmed Mansour for conducting an in-depth review of the paper. Our gratitude also extends to Proteinea's deep learning and bioinformatics teams, whose provision of hardware, software, and broad project support was invaluable. It is important to note that this research received no specific grant from any funding agency in the public, commercial, or not-for-profit sectors.
\end{ack}

\bibliographystyle{unsrtnat}
\bibliography{main}

\clearpage
\appendix
\section{Technical appendices and supplementary material}
\subsection{Full results}
{\small\tabcolsep=3.0pt \begin{longtable}{ll ccc ccc}
    \caption{Table presents Spearman ($\rho$), Pearson (r) correlations, and Root Mean Squared Error (RMSE) in $pK_d$ units for validation and test sets. Metrics are derived from the best performing checkpoint (selected based on highest validation $\rho$) for each setup (PLM/method/architecture combination) and are reported as mean $\pm$ standard deviation over three independent runs with different seeds. Setup column abbreviations: CV: ConvBERT; FT: Finetuning; PAD: Pooled attention addition; HP: Hierarchical pooling; SC: Sequences concatenation; EC: Embeddings concatenation}
    \label{tab_full}\\
    \toprule
    \multirow{2}{*}{\textbf{PLM}} & \multirow{2}{*}{\textbf{Setup}} &
    \multicolumn{3}{c}{\textbf{Validation Split}} & \multicolumn{3}{c}{\textbf{Test Split}} \\
    \cmidrule(lr){3-5}\cmidrule(lr){6-8}
     & & \textbf{Spearman} & \textbf{Pearson} & \textbf{RMSE} & \textbf{Spearman} & \textbf{Pearson} & \textbf{RMSE} \\
    \midrule
    \endhead

    \hline
    \multicolumn{8}{r}{\textit{Continued on next page}} \\
    \hline
    \endfoot

    \hline
    \endlastfoot
    ProtT5 & CV-PAD & 0.48 $\pm$ 0.02 & 0.48 $\pm$ 0.01 & 1.52 $\pm$ 0.09 & 0.48 $\pm$ 0.03 & 0.51 $\pm$ 0.02 & 1.42 $\pm$ 0.10 \\
    Ankh2-Ext1 & FT-HP & 0.48 $\pm$ 0.01 & 0.49 $\pm$ 0.01 & 1.50 $\pm$ 0.10 & 0.47 $\pm$ 0.01 & 0.48 $\pm$ 0.01 & 1.45 $\pm$ 0.13 \\
    ESM2-650M & CV-HP & 0.44 $\pm$ 0.03 & 0.43 $\pm$ 0.03 & 1.86 $\pm$ 0.29 & 0.47 $\pm$ 0.02 & 0.48 $\pm$ 0.02 & 1.74 $\pm$ 0.33 \\
    Ankh2-Ext2 & FT-HP & 0.47 $\pm$ 0.01 & 0.47 $\pm$ 0.01 & 1.51 $\pm$ 0.07 & 0.45 $\pm$ 0.01 & 0.46 $\pm$ 0.02 & 1.43 $\pm$ 0.06 \\
    ESM2-3B & FT-PAD & 0.45 $\pm$ 0.01 & 0.45 $\pm$ 0.01 & 1.49 $\pm$ 0.07 & 0.44 $\pm$ 0.01 & 0.46 $\pm$ 0.00 & 1.43 $\pm$ 0.08 \\
    ESM2-650M & CV-PAD & 0.44 $\pm$ 0.00 & 0.44 $\pm$ 0.00 & 1.57 $\pm$ 0.12 & 0.44 $\pm$ 0.01 & 0.46 $\pm$ 0.01 & 1.46 $\pm$ 0.10 \\
    Ankh-Base & FT-HP & 0.47 $\pm$ 0.01 & 0.47 $\pm$ 0.00 & 1.47 $\pm$ 0.03 & 0.44 $\pm$ 0.01 & 0.45 $\pm$ 0.01 & 1.41 $\pm$ 0.03 \\
    ESM2-650M & FT-HP & 0.45 $\pm$ 0.02 & 0.44 $\pm$ 0.02 & 1.75 $\pm$ 0.29 & 0.44 $\pm$ 0.02 & 0.45 $\pm$ 0.01 & 1.68 $\pm$ 0.29 \\
    ESM2-3B & FT-SC & 0.48 $\pm$ 0.01 & 0.47 $\pm$ 0.01 & 1.42 $\pm$ 0.01 & 0.44 $\pm$ 0.01 & 0.47 $\pm$ 0.01 & 1.34 $\pm$ 0.00 \\
    ProtT5 & FT-HP & 0.47 $\pm$ 0.02 & 0.46 $\pm$ 0.02 & 1.51 $\pm$ 0.05 & 0.44 $\pm$ 0.01 & 0.45 $\pm$ 0.01 & 1.47 $\pm$ 0.08 \\
    ProtT5 & CV-HP & 0.42 $\pm$ 0.02 & 0.42 $\pm$ 0.02 & 1.66 $\pm$ 0.23 & 0.44 $\pm$ 0.01 & 0.44 $\pm$ 0.01 & 1.59 $\pm$ 0.25 \\
    ProtT5 & FT-PAD & 0.44 $\pm$ 0.00 & 0.44 $\pm$ 0.00 & 1.65 $\pm$ 0.13 & 0.44 $\pm$ 0.01 & 0.45 $\pm$ 0.01 & 1.57 $\pm$ 0.14 \\
    Ankh2-Ext1 & FT-PAD & 0.46 $\pm$ 0.02 & 0.46 $\pm$ 0.02 & 1.48 $\pm$ 0.01 & 0.43 $\pm$ 0.00 & 0.44 $\pm$ 0.00 & 1.40 $\pm$ 0.02 \\
    Ankh2-Ext2 & FT-PAD & 0.47 $\pm$ 0.01 & 0.47 $\pm$ 0.01 & 1.49 $\pm$ 0.08 & 0.43 $\pm$ 0.01 & 0.44 $\pm$ 0.01 & 1.43 $\pm$ 0.09 \\
    ESM2-650M & FT-SC & 0.47 $\pm$ 0.03 & 0.47 $\pm$ 0.03 & 1.49 $\pm$ 0.01 & 0.43 $\pm$ 0.03 & 0.46 $\pm$ 0.02 & 1.43 $\pm$ 0.03 \\
    Ankh-Large & FT-PAD & 0.45 $\pm$ 0.00 & 0.45 $\pm$ 0.01 & 1.56 $\pm$ 0.12 & 0.43 $\pm$ 0.04 & 0.45 $\pm$ 0.04 & 1.46 $\pm$ 0.13 \\
    ProtT5 & CV-EC & 0.40 $\pm$ 0.01 & 0.40 $\pm$ 0.01 & 1.56 $\pm$ 0.00 & 0.43 $\pm$ 0.03 & 0.44 $\pm$ 0.03 & 1.46 $\pm$ 0.00 \\
    ESM2-3B & CV-HP & 0.44 $\pm$ 0.02 & 0.43 $\pm$ 0.02 & 1.58 $\pm$ 0.03 & 0.43 $\pm$ 0.03 & 0.44 $\pm$ 0.03 & 1.50 $\pm$ 0.01 \\
    Ankh-Base & FT-PAD & 0.47 $\pm$ 0.01 & 0.47 $\pm$ 0.01 & 1.48 $\pm$ 0.04 & 0.42 $\pm$ 0.02 & 0.44 $\pm$ 0.01 & 1.43 $\pm$ 0.03 \\
    Ankh-Base & CV-PAD & 0.46 $\pm$ 0.01 & 0.45 $\pm$ 0.01 & 1.54 $\pm$ 0.12 & 0.42 $\pm$ 0.01 & 0.44 $\pm$ 0.01 & 1.47 $\pm$ 0.12 \\
    ESM2-3B & FT-HP & 0.44 $\pm$ 0.01 & 0.43 $\pm$ 0.01 & 1.51 $\pm$ 0.07 & 0.41 $\pm$ 0.02 & 0.44 $\pm$ 0.02 & 1.46 $\pm$ 0.09 \\
    ProtT5 & CV-SC & 0.44 $\pm$ 0.01 & 0.44 $\pm$ 0.01 & 1.51 $\pm$ 0.03 & 0.41 $\pm$ 0.00 & 0.43 $\pm$ 0.00 & 1.44 $\pm$ 0.03 \\
    ESM2-3B & FT-EC & 0.45 $\pm$ 0.02 & 0.45 $\pm$ 0.03 & 1.50 $\pm$ 0.08 & 0.41 $\pm$ 0.02 & 0.44 $\pm$ 0.02 & 1.44 $\pm$ 0.05 \\
    ESM2-650M & CV-EC & 0.47 $\pm$ 0.02 & 0.46 $\pm$ 0.02 & 1.52 $\pm$ 0.02 & 0.41 $\pm$ 0.01 & 0.43 $\pm$ 0.00 & 1.48 $\pm$ 0.02 \\
    Ankh-Large & CV-PAD & 0.44 $\pm$ 0.03 & 0.43 $\pm$ 0.03 & 1.75 $\pm$ 0.30 & 0.41 $\pm$ 0.02 & 0.44 $\pm$ 0.02 & 1.68 $\pm$ 0.36 \\
    Ankh2-Ext1 & CV-PAD & 0.46 $\pm$ 0.01 & 0.45 $\pm$ 0.01 & 1.73 $\pm$ 0.43 & 0.41 $\pm$ 0.02 & 0.43 $\pm$ 0.02 & 1.67 $\pm$ 0.41 \\
    Ankh2-Ext2 & CV-SC & 0.44 $\pm$ 0.03 & 0.44 $\pm$ 0.03 & 1.54 $\pm$ 0.04 & 0.41 $\pm$ 0.02 & 0.41 $\pm$ 0.02 & 1.49 $\pm$ 0.02 \\
    Ankh2-Ext1 & CV-SC & 0.44 $\pm$ 0.02 & 0.43 $\pm$ 0.02 & 1.59 $\pm$ 0.13 & 0.41 $\pm$ 0.02 & 0.41 $\pm$ 0.02 & 1.54 $\pm$ 0.15 \\
    Ankh-Base & CV-HP & 0.40 $\pm$ 0.01 & 0.40 $\pm$ 0.00 & 1.67 $\pm$ 0.14 & 0.41 $\pm$ 0.03 & 0.43 $\pm$ 0.04 & 1.61 $\pm$ 0.16 \\
    ProtT5 & FT-EC & 0.45 $\pm$ 0.01 & 0.44 $\pm$ 0.01 & 1.60 $\pm$ 0.17 & 0.41 $\pm$ 0.03 & 0.41 $\pm$ 0.02 & 1.58 $\pm$ 0.20 \\
    Ankh-Large & CV-HP & 0.42 $\pm$ 0.01 & 0.41 $\pm$ 0.01 & 1.80 $\pm$ 0.23 & 0.40 $\pm$ 0.03 & 0.40 $\pm$ 0.03 & 1.78 $\pm$ 0.30 \\
    Ankh2-Ext2 & CV-PAD & 0.44 $\pm$ 0.01 & 0.44 $\pm$ 0.02 & 1.57 $\pm$ 0.06 & 0.40 $\pm$ 0.01 & 0.43 $\pm$ 0.02 & 1.51 $\pm$ 0.05 \\
    ProtT5 & FT-SC & 0.46 $\pm$ 0.03 & 0.46 $\pm$ 0.03 & 1.53 $\pm$ 0.02 & 0.40 $\pm$ 0.01 & 0.43 $\pm$ 0.00 & 1.51 $\pm$ 0.01 \\
    ESM2-3B & CV-EC & 0.45 $\pm$ 0.00 & 0.45 $\pm$ 0.00 & 1.69 $\pm$ 0.09 & 0.40 $\pm$ 0.01 & 0.42 $\pm$ 0.01 & 1.65 $\pm$ 0.11 \\
    Ankh-Base & CV-SC & 0.42 $\pm$ 0.03 & 0.43 $\pm$ 0.03 & 1.56 $\pm$ 0.07 & 0.39 $\pm$ 0.04 & 0.41 $\pm$ 0.04 & 1.55 $\pm$ 0.09 \\
    ESM2-3B & CV-SC & 0.42 $\pm$ 0.00 & 0.42 $\pm$ 0.00 & 1.84 $\pm$ 0.15 & 0.39 $\pm$ 0.03 & 0.42 $\pm$ 0.03 & 1.75 $\pm$ 0.12 \\
    ESM2-650M & CV-SC & 0.42 $\pm$ 0.03 & 0.42 $\pm$ 0.02 & 1.56 $\pm$ 0.04 & 0.39 $\pm$ 0.01 & 0.41 $\pm$ 0.01 & 1.51 $\pm$ 0.04 \\
    ESM2-650M & FT-PAD & 0.43 $\pm$ 0.01 & 0.43 $\pm$ 0.02 & 1.56 $\pm$ 0.08 & 0.39 $\pm$ 0.02 & 0.39 $\pm$ 0.01 & 1.49 $\pm$ 0.04 \\
    ESM2-650M & FT-EC & 0.45 $\pm$ 0.03 & 0.44 $\pm$ 0.03 & 1.51 $\pm$ 0.02 & 0.39 $\pm$ 0.03 & 0.41 $\pm$ 0.04 & 1.46 $\pm$ 0.06 \\
    Ankh2-Ext2 & FT-EC & 0.41 $\pm$ 0.00 & 0.41 $\pm$ 0.00 & 1.51 $\pm$ 0.01 & 0.39 $\pm$ 0.06 & 0.40 $\pm$ 0.05 & 1.45 $\pm$ 0.05 \\
    Ankh2-Ext2 & CV-HP & 0.43 $\pm$ 0.03 & 0.41 $\pm$ 0.03 & 1.86 $\pm$ 0.47 & 0.39 $\pm$ 0.03 & 0.39 $\pm$ 0.03 & 1.82 $\pm$ 0.51 \\
    Ankh2-Ext1 & CV-HP & 0.42 $\pm$ 0.02 & 0.41 $\pm$ 0.02 & 1.70 $\pm$ 0.03 & 0.39 $\pm$ 0.02 & 0.40 $\pm$ 0.02 & 1.66 $\pm$ 0.07 \\
    Ankh-Large & FT-HP & 0.42 $\pm$ 0.01 & 0.42 $\pm$ 0.01 & 1.52 $\pm$ 0.04 & 0.38 $\pm$ 0.02 & 0.40 $\pm$ 0.03 & 1.48 $\pm$ 0.06 \\
    Ankh2-Ext2 & CV-EC & 0.44 $\pm$ 0.02 & 0.44 $\pm$ 0.02 & 1.56 $\pm$ 0.05 & 0.38 $\pm$ 0.02 & 0.39 $\pm$ 0.02 & 1.53 $\pm$ 0.00 \\
    Ankh2-Ext1 & FT-SC & 0.41 $\pm$ 0.02 & 0.42 $\pm$ 0.02 & 1.56 $\pm$ 0.05 & 0.38 $\pm$ 0.03 & 0.40 $\pm$ 0.03 & 1.50 $\pm$ 0.05 \\
    Ankh-Large & CV-SC & 0.43 $\pm$ 0.01 & 0.43 $\pm$ 0.01 & 1.72 $\pm$ 0.19 & 0.38 $\pm$ 0.03 & 0.38 $\pm$ 0.03 & 1.69 $\pm$ 0.17 \\
    Ankh-Base & CV-EC & 0.42 $\pm$ 0.01 & 0.42 $\pm$ 0.01 & 1.59 $\pm$ 0.11 & 0.38 $\pm$ 0.04 & 0.40 $\pm$ 0.04 & 1.58 $\pm$ 0.19 \\
    ESM2-3B & CV-PAD & 0.44 $\pm$ 0.03 & 0.43 $\pm$ 0.03 & 2.82 $\pm$ 1.18 & 0.38 $\pm$ 0.04 & 0.40 $\pm$ 0.05 & 2.75 $\pm$ 1.15 \\
    ESM3-SM-Open & FT-PAD & 0.42 $\pm$ 0.03 & 0.42 $\pm$ 0.02 & 1.54 $\pm$ 0.10 & 0.37 $\pm$ 0.01 & 0.38 $\pm$ 0.01 & 1.49 $\pm$ 0.08 \\
    Ankh-Large & CV-EC & 0.44 $\pm$ 0.01 & 0.43 $\pm$ 0.01 & 1.53 $\pm$ 0.06 & 0.37 $\pm$ 0.02 & 0.38 $\pm$ 0.01 & 1.50 $\pm$ 0.07 \\
    ESM3-SM-Open & CV-EC & 0.44 $\pm$ 0.01 & 0.44 $\pm$ 0.01 & 1.73 $\pm$ 0.23 & 0.36 $\pm$ 0.01 & 0.37 $\pm$ 0.02 & 1.77 $\pm$ 0.23 \\
    ESM3-SM-Open & CV-PAD & 0.43 $\pm$ 0.02 & 0.42 $\pm$ 0.02 & 1.81 $\pm$ 0.19 & 0.36 $\pm$ 0.04 & 0.38 $\pm$ 0.03 & 1.80 $\pm$ 0.25 \\
    ESM3-SM-Open & FT-HP & 0.41 $\pm$ 0.03 & 0.41 $\pm$ 0.03 & 1.56 $\pm$ 0.02 & 0.36 $\pm$ 0.03 & 0.36 $\pm$ 0.03 & 1.53 $\pm$ 0.01 \\
    ESM3-SM-Open & FT-EC & 0.37 $\pm$ 0.04 & 0.38 $\pm$ 0.04 & 1.57 $\pm$ 0.07 & 0.36 $\pm$ 0.05 & 0.38 $\pm$ 0.05 & 1.48 $\pm$ 0.05 \\
    Ankh2-Ext1 & CV-EC & 0.44 $\pm$ 0.02 & 0.43 $\pm$ 0.01 & 1.69 $\pm$ 0.08 & 0.36 $\pm$ 0.05 & 0.38 $\pm$ 0.05 & 1.69 $\pm$ 0.10 \\
    Ankh2-Ext2 & FT-SC & 0.42 $\pm$ 0.02 & 0.43 $\pm$ 0.02 & 1.50 $\pm$ 0.03 & 0.35 $\pm$ 0.02 & 0.39 $\pm$ 0.03 & 1.47 $\pm$ 0.05 \\
    Ankh-Large & FT-EC & 0.39 $\pm$ 0.00 & 0.40 $\pm$ 0.01 & 1.52 $\pm$ 0.02 & 0.35 $\pm$ 0.05 & 0.39 $\pm$ 0.04 & 1.46 $\pm$ 0.06 \\
    ESM3-SM-Open & CV-HP & 0.38 $\pm$ 0.00 & 0.38 $\pm$ 0.00 & 1.54 $\pm$ 0.02 & 0.35 $\pm$ 0.01 & 0.36 $\pm$ 0.00 & 1.49 $\pm$ 0.03 \\
    ESM3-SM-Open & CV-SC & 0.44 $\pm$ 0.03 & 0.42 $\pm$ 0.03 & 1.81 $\pm$ 0.27 & 0.35 $\pm$ 0.02 & 0.36 $\pm$ 0.02 & 1.82 $\pm$ 0.27 \\
    Ankh-Large & FT-SC & 0.42 $\pm$ 0.01 & 0.43 $\pm$ 0.01 & 1.51 $\pm$ 0.05 & 0.34 $\pm$ 0.01 & 0.39 $\pm$ 0.02 & 1.47 $\pm$ 0.05 \\
    Ankh-Base & FT-EC & 0.40 $\pm$ 0.01 & 0.41 $\pm$ 0.01 & 1.52 $\pm$ 0.03 & 0.33 $\pm$ 0.08 & 0.34 $\pm$ 0.06 & 1.52 $\pm$ 0.06 \\
    Ankh-Base & FT-SC & 0.38 $\pm$ 0.05 & 0.39 $\pm$ 0.05 & 1.54 $\pm$ 0.03 & 0.32 $\pm$ 0.01 & 0.37 $\pm$ 0.01 & 1.51 $\pm$ 0.05 \\
    Ankh2-Ext1 & FT-EC & 0.40 $\pm$ 0.01 & 0.40 $\pm$ 0.01 & 1.55 $\pm$ 0.01 & 0.32 $\pm$ 0.06 & 0.36 $\pm$ 0.05 & 1.54 $\pm$ 0.04 \\
    ESM3-SM-Open & FT-SC & 0.27 $\pm$ 0.01 & 0.27 $\pm$ 0.01 & 6.50 $\pm$ 0.08 & 0.28 $\pm$ 0.01 & 0.28 $\pm$ 0.02 & 6.31 $\pm$ 0.09 \\
    \bottomrule
\end{longtable}}

\newpage

\subsection{Additional Methods Figures}
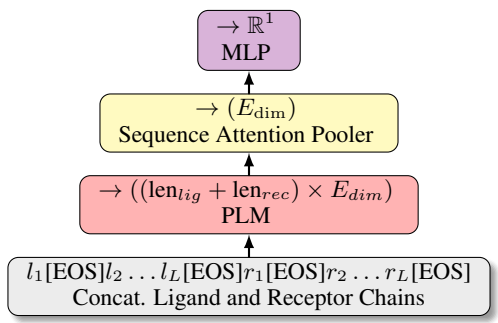
\begin{figure*}[h!]
    \centering
        \begin{tikzpicture}[
            font=\footnotesize,
            >=latex,
            line join=round,
            line cap=round,
            node distance=0.3cm and 0.3cm,
            base_box/.style={
                draw,
                rounded corners,
                inner sep=1.0pt,
                minimum width=0.0cm,
                minimum height=0.1cm,
                align=center
            },
            input_box/.style={ 
                base_box,
                blur shadow,
                shape=rectangle,
                rotate=0,
                fill=gray!15
            },
            plm_box/.style={ 
                base_box,
                fill=red!30
            },
            pool1_box/.style={ 
                base_box,
                fill=yellow!30
            },
            pool2_box/.style={ 
                base_box,
                fill=yellow!30
            },
            concat_box/.style={ 
                base_box,
                fill=orange!30
            },
            mlp_box/.style={ 
                base_box,
                fill=violet!30
            }
        ]
        
        
        \node[input_box] (lig_in) {\begin{tabular}{c}
        \(l_1\text{[EOS]}l_2\dots l_L\text{[EOS]}r_1\text{[EOS]}r_2\dots r_L\text{[EOS]}\) \\ 
        Concat. Ligand and Receptor Chains 
        \end{tabular}};
        
        \node[plm_box, above=of lig_in] (lig_plm) {
        \begin{tabular}{c}
        \(\rightarrow ((\text{len}_{lig}+\text{len}_{rec})\times E_{dim})\) \\ 
        PLM 
        \end{tabular}
        };
        \draw[->, thick] (lig_in) -- (lig_plm);
        
        \node[pool2_box, above=of lig_plm] (lig_pool_chain) {
        \begin{tabular}{c}
        \(\rightarrow (E_{\dim})\)\\ 
        Sequence Attention Pooler 
        \end{tabular}
        };
        \draw[->, thick] (lig_plm) -- (lig_pool_chain);

                
        \node[mlp_box, above=of lig_pool_chain] (mlp) {
        \begin{tabular}{c}
        \(\rightarrow \mathbb{R}^1\)\\ 
        MLP 
        \end{tabular}
        };
        \draw[->, thick] (lig_pool_chain) -- (mlp);
        
        \end{tikzpicture}
    \caption{Sequences Concatenation Architecture}
    \label{fig:sc}
\end{figure*}

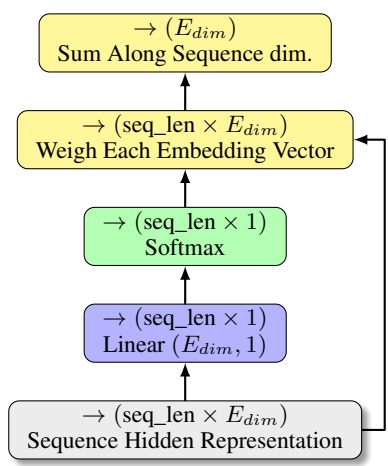
\begin{figure*}[h!]
    \centering
        \begin{tikzpicture}[
            font=\footnotesize,
            >=latex,
            line join=round,
            line cap=round,
            node distance=0.5cm and 0.3cm,
            base_box/.style={
                draw,
                rounded corners,
                inner sep=1.0pt,
                minimum width=0.0cm,
                minimum height=0.1cm,
                align=center
            },
            input_box/.style={ 
                base_box,
                blur shadow,
                shape=rectangle,
                rotate=0,
                fill=gray!15
            },
            linear_box/.style={ 
                base_box,
                fill=blue!30
            },
            operation_box/.style={ 
                base_box,
                fill=yellow!50
            },
            softmax_box/.style={ 
                base_box,
                fill=green!30
            }
        ]
        
        
        \node[input_box] (in) {\begin{tabular}{c}
        \(\rightarrow (\text{seq\_len}\times E_{dim})\) \\
        Sequence Hidden Representation 
        \end{tabular}};
        
        \node[linear_box, above=of in] (linear) {\begin{tabular}{c}
        \(\rightarrow (\text{seq\_len}\times 1)\)\\
        Linear $(E_{dim}, 1)$ 
        \end{tabular}};
        \draw[->, thick] (in) -- (linear);

        \node[softmax_box, above=of linear] (softmax) {\begin{tabular}{c}
        \(\rightarrow (\text{seq\_len}\times 1)\)\\
        Softmax 
        \end{tabular}};
        \draw[->, thick] (linear) -- (softmax);

        \node[operation_box, above=of softmax] (mult) {\begin{tabular}{c}
        \(\rightarrow (\text{seq\_len}\times E_{dim})\) \\
        Weigh Each Embedding Vector 
        \end{tabular}};
        \draw[->, thick] (softmax) -- (mult);

        \coordinate (in_r) at ($(in.east)+(0.3cm,0)$); 
        \coordinate (mult_r) at (in_r |- mult.east);        
        \draw[->, thick] (in.east) -- (in_r) -- (mult_r) -- (mult.east); 

        \node[operation_box, above=of mult] (sum) {\begin{tabular}{c}
        \(\rightarrow (E_{dim})\) \\
        Sum Along Sequence dim. 
        \end{tabular}};
        \draw[->, thick] (mult) -- (sum);
        \end{tikzpicture}
    \caption{Global $1D$ Attention Pooler Architecture. A linear layer transforms the input sequence of hidden states (each of dimension $E_{dim}$) into a vector of scalar attention scores, one per hidden state. These scores are subsequently normalized via a softmax function to produce attention weights. The final pooled output, a single vector of dimension $E_{dim}$, is computed as the weighted sum of the original hidden states using these attention weights. Dimensions displayed on each block denote the output dimensions of that component}
    \label{fig:attn_poolin}
\end{figure*}

\end{document}